# PyTorch Tabular: A Framework for Deep Learning with Tabular Data

Manu Joseph

*Abstract* — In spite of showing unreasonable effectiveness in modalities like Text and Image, Deep Learning has always lagged Gradient Boosting in tabular data—both in popularity and performance. But recently there have been newer models created specifically for tabular data, which is pushing the performance bar. But popularity is still a challenge because there is no easy, ready-to-use library like Sci-Kit Learn for deep learning. PyTorch Tabular is a new deep learning library which makes working with Deep Learning and tabular data easy and fast. It is a library built on top of PyTorch and PyTorch Lightning and works on pandas dataframes directly. Many SOTA models like NODE and TabNet are already integrated and implemented in the library with a unified API. PyTorch Tabular is designed to be easily extensible for researchers, simple for practitioners, and robust in industrial deployments. The library is available at https://github.com/manujosephv/pytorch_tabular

## I. INTRODUCTION

The unreasonable effectiveness of Deep Learning that was displayed in many other modalities—like text[1] and image[2]—have not been thoroughly demonstrated in tabular data. Despite being the most used data type in real-world problems, tabular modality is relatively less explored in Deep Learning literature. The state-of-the-art performance in many problems with tabular data is often achieved by "shallow" models, such as gradient boosted decision trees (GBDT)[3] (XGBoost[4], LightGBM[5], CatBoost[6]). If "*performance*" is one dimension along which GBDTs beat Deep Learning approaches, "*popularity*" is another. If we look at the different machine learning competitions (e.g. Kaggle), we can see the popularity of GBDTs, which are almost always part of the winning solutions.

The past few years, we have seen an increased interest in this modality and many works address this modality to push the state-of-the-art. Deep Forest[7], TabNN[8], TabNet[9], Neural Oblivious Decision Ensembles[10] are just a few architectures proposed specifically for tabular modality. Out of these, NODE and TabNet has shown to beat the GBDT baselines as well.

While research has started to push the "*performance*" bar on tabular data, the "*popularity*" bar is still low. One of the primary reasons behind this is the lack of support on the software side of things. Training Deep Learning models are still and involved process with quite a bit of software engineering required. When this is compared to the ease Scikit-learn[11] and other libraries adopting the Scikit-learn API provides practitioners, we get a clue as to why popularity of Deep Learning for Tabular data is still low.

PyTorch Tabular is a library which aims to make Deep Learning with Tabular data easy and accessible to real-world cases and research alike. The core principles behind the design of the library are:

- Low Resistance Usability
- Easy Customization
- Scalable and Easier to Deploy

PyTorch Tabular attempts to make the "software engineering" part of working with Neural Networks as easy and effortless as possible and let you focus on the model. It also hopes to unify the different developments in the Tabular space into a single framework with a unified API that will work with different state-of-the-art models. It also provides an easily extensible BaseModel to aid Deep Learning researchers create new architectures for tabular data.

PyTorch Tabular is built on the shoulders of giants like PyTorch[12], PyTorch Lightning[13], and Pandas[14]. The library is released under the MIT license and is available on GitHub[1]. Detailed documentation and tutorials are available on documentation page[2].

## II. RELATED WORK

The ML community has a strong culture of building open-source tools, which both accelerated research and adoption of new techniques in the industry. Tensorflow[15], PyTorch, and similar frameworks started the journey of abstraction of Deep Learning implementation by providing automatic differentiation, pre-built fundamental blocks of Neural Networks etc. PyTorch Lightning came in and abstracted away the training loop for PyTorch and enabled easy and scalable training. PyTorch Tabular takes that journey of abstraction to the next level by providing domain specific abstraction layer.

The concept of having domain specific abstractions to train neural network models originated from fastai[16] and fastai.tabular provides an easy-to-use API for training deep learning models for tabular data. But where PyTorch Tabular is different is in the fact that it is a strongly de-coupled implementation and relies on standard components like Base PyTorch layers, optimizers and loss functions. The training loop is handled by Pytorch Lightning, which is also growing to be a standard in the community. This makes PyTorch Tabular much more extensible for custom use cases.

---

[1] https://github.com/manujosephv/pytorch_tabular
[2] https://pytorch-tabular.readthedocs.io/en/latest/

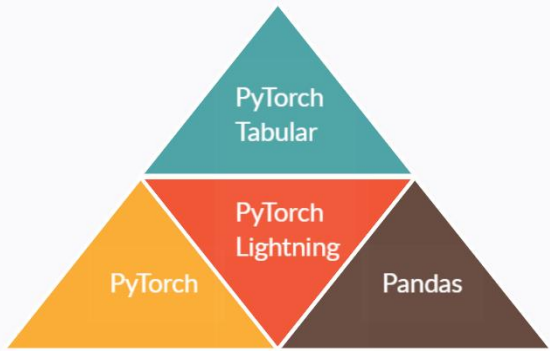

Fig. 1. PyTorch Tabular is built on strong foundations of tried and tested frameworks.

III. LIBRARY DESIGN

PyTorch Tabular is designed to make the standard modelling pipeline easy enough for practitioners as well as standard enough for production deployment. In addition to that, it also has a focus on customization to enable wide usage in research.

Pytorch Tabular has adopted a 'config-driven' approach to satisfy these objectives.

A. Config Driven

There are 5 config files which drives the whole process-

1. *DataConfig* – *DataConfig* is where you define the parameters regarding how you manage data within your pipeline. We distinguish between categorical and continuous features, decide the normalization, or feature transformations, etc. in this config.
2. *ModelConfig* – There is a separate *ModelConfig* defined for each model that is implemented in PyTorch Tabular. It inherits from a base ModelConfig which holds common parameters like *task (classification or regression), learning_rate, loss, metrics,* etc. Each model that is implemented inherits these parameters and adds model specific hyperparameters to the config. By choosing the corresponding *ModelConfig*, Pytorch Tabular automatically initializes the right model.
3. *TrainerConfig*—*TrainerConfig* handles all the parameters to control your training process, and most of these parameters are passed down to the PyTorch Lightning layer. You can set parameters like *batch_size, max_epochs, early_stopping*, etc. in here.
4. *OptimizerConfig*—Optimizers and Learning Rate Schedulers are another integral part in training a neural network. These configurations can be done using the OptimizerConfig.
5. *ExperimentConfig* - Experiment Tracking is almost an essential part of machine learning. It is critical in upholding reproducibility. PyTorch Tabular embraces this and supports experiment tracking internally. Currently, PyTorch Tabular supports two experiment Tracking Framework—Tensorboard and Weights & Biases.

Tensorboard logging is barebones. PyTorch Tabular just logs the losses and metrics to Tensorboard. W&B tracking is much more feature rich - in addition to tracking losses and metrics, it can also track the gradients of the different layers, logits of your model across epochs, etc.

These config files can be set programmatically as well as through YAML files, which makes this easy for both Data Scientists and ML Engineers.

B. BaseModel

PyTorch Tabular uses an abstract class—BaseModel—which implements the standard part of any model definition like loss and metric calculation, etc. This class serves as a template on which any other model is implemented and ensures smooth interoperability between the model and the training engine. Inheriting this class, the only two methods that a new model must implement are the model initialization part and the forward pass. And in case you need to do something non-standard in the loss calculation, all you have to do is overwrite the corresponding methods in your model definition.

C. Data Module

PyTorch Tabular uses Data Module, as defined by Pytorch Lightning, to unify and standardize the data processing. It encompasses the preprocessing, label encoding, categorical encoding, feature transformations, target transformations, etc. and also ensures the same data processing is applied to train and validation splits, as well as new and unseen data. It provides PyTorch dataloaders for training and inference.

D. TabularModel

*TabularModel* is the core component which brings together the configs, initializes the right model, the data module, and handles the train and prediction functions with methods like '*fit*' and '*predict*'.

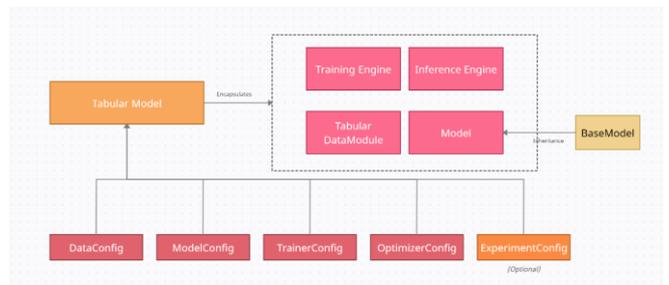

Fig. 2. Overall Structure of PyTorch Tabular

IV. IMPLEMENTED MODELS AND UNIFIED API

PyTorch Tabular has implemented a few state-of-the-art model architectures and unified them with a single, easy-to-use API which is well suited for rapid iterations. For Deep Learning to gain popularity among practitioners, it is important to be able to provide an easily used API which can compare to the ease of use given by the "Scikit-learn" APIs.

The models which are currently implemented in PyTorch Tabular are:

A. CategoryEmbeddingModel

This is a standard feed-forward network with the categorical features passed through a learnable embedding layer. The model architecture is very similar to the Tabular model in fastai with BatchNorm and Dropout Layers in between standard linear layers.

*B. Neural Oblivious Decision Ensembles (NODE)*

NODE[10] is a model architecture presented in ICLR 2020 and shown to beat tuned GBDT models on several datasets. It uses a Neural equivalent of Oblivious Trees (the kind of trees CatBoost[6] uses) as the basic building blocks of the architecture.

There are two variants of this algorithm implemented in PyTorch Tabular—*NODEModel* and *CategoryEmbeddingNODEModel*. The only difference is in the way categorical features are treated. In *NODEModel*, the categorical features are encoded using LeaveOneOutEncoding[17] (as suggested by the authors) and in *CategoryEmbeddingNODEModel* the categorical embeddings are learned from data.

*C. TabNet*

TabNet[9] is a model architecture which deviates from the tree-based hybrid design philosophy and uses Sparse Attention in multiple steps of decision making to model the output. The architecture consists of sequential learnable decision steps which includes feature selection using a learnable mask. The multiple steps create higher representations of the input data which is used for the final task.

*D. AutoInt*

AutoInt[18] is a model architecture, first proposed for Click-through rate prediction. This architecture tries to handle sparse features efficiently and automatically learn cross-features or interactions between the features using an attention mechanism.

*E. Usage*

The basic usage is fairly simple. Below is an example. We define the configs and select the *CategoryEmbeddingModelConfig* as the model config. All the parameters have intelligent defaults so that you can get started as soon as possible.

```
data_config = DataConfig(
    target=['target'],
    continuous_cols=num_col_names,
    categorical_cols=cat_col_names,
)
trainer_config = TrainerConfig(
    gpus=1, #index of the GPU to use. 0, means CPU
)
optimizer_config = OptimizerConfig()

model_config = CategoryEmbeddingModelConfig(
    task="classification"
)
experiment_config = ExperimentConfig(
    project_name="PyTorch Tabular Example"
)
```

Now that the configs are defined, we can put it all together in a *TabularModel*.

```
tabular_model = TabularModel(
    data_config=data_config,
    model_config=model_config,
    optimizer_config=optimizer_config,
    trainer_config=trainer_config,
)
```

The TabularModel takes in the configs and sets up the whole modelling pipeline. Now We just need to call the *fit* method and pass the train and test dataframes. We can also pass in validation dataframe. But if omitted, *TabularModel* will separate 20% (also configurable) at random from the data as validation.

By default, *EarlyStopping* is enabled and is monitoring validation loss with a patience of 3 epochs. The trainer also saves the best model (based on validation loss) and loads that model at the end of training. *TrainerConfig* has the parameters to tweak this default behaviour.

```
tabular_model.fit(train=train, validation=val)
```

After the training, there are three actions that you usually take in a typical modelling pipeline.

1. Evaluate the model on some new data

```
result = tabular_model.evaluate(test)
```

2. Get predictions on new data

```
pred_df = tabular_model.predict(test)
```

3. Save and Load the Model

```
tabular_model.save_model("examples/basic")
loaded_model = TabularModel.load_from_checkpoint(
    "examples/basic"
)
```

Detailed documentation and tutorials for common tasks are present in the documentation.

## V. CONCLUSION

Deep Learning for tabular data is gaining popularity in the research community as well as the industry and in the face of growing popularity, it is essential to have a unified and easy to use API for tabular data, similar to what scikit-learn has done for classical machine learning algorithms. PyTorch Tabular is hoping to fill in that space and reduce the barrier for entry in using new state-of-the-art deep learning model architectures in industry use cases. It also hopes to reduce the "engineering" work for researchers who are working on new model architectures.

## VI. FUTURE WORK

PyTorch Tabular is a relatively new library and will continue to grow. We actively invite contributors to help maintain and grow the library. Future roadmap is available on the Github Readme. The items are along the below three paradigms:

1. Adding new models
2. Integrating Hyperparameter Tuning
3. Adding Text and Image modalities for multi-modal problems.
4. Adding new preprocessing techniques
5. Adding Self-Supervised learning architectures